\documentclass[runningheads]{llncs}
\usepackage[T1]{fontenc}
\usepackage{amsmath}
\usepackage{amssymb}
\usepackage{tikz}
\usepackage{longtable}
\usepackage{amsmath,amssymb,amsfonts}
\usepackage{algorithmic}
\usepackage{graphicx}
\usepackage{textcomp}
\usepackage{orcidlink}
\usepackage{wrapfig}
\usepackage{graphicx,verbatim}
\usepackage{graphicx}
\usepackage{amsmath}
\usepackage{amssymb}
\usepackage{booktabs}
\usepackage[dvipsnames]{xcolor}
\usepackage{xcolor,colortbl}
\usepackage{color, colortbl}
\usepackage{xcolor}

\usepackage{sidecap}

\usepackage{float}
\definecolor{purple}{RGB}{230, 227, 254}
\definecolor{lightgreen}{RGB}{238, 252, 241}
\definecolor{lightred}{RGB}{231, 187, 187}
\definecolor{darkred}{RGB}{198, 129, 129}

\definecolor{tabhighlight}{HTML}{e5e5e5}

\usepackage{graphicx, amsmath, amssymb, caption, subcaption, multirow, overpic}
\definecolor{tabhighlightcolor2}{HTML}{e5e5e5}
\definecolor{citecolor}{HTML}{0071bc}

\usepackage{graphicx,verbatim}
\begin{document}
\title{Decoupling Clinical and Class-Agnostic Features for Reliable Few-Shot Adaptation under Shift}
\titlerunning{Decoupling Clinical and Class-Agnostic Features under Shift}

\author{\small{Umaima Rahman\inst{1} \and Raza Imam\inst{1} \and Mohammad Yaqub\inst{1} \and Dwarikanath Mahapatra\inst{2}} }

\institute{\small {Mohamed Bin Zayed University of Artificial Intelligence, Abu Dhabi, UAE\\
\and
Khalifa University, Abu Dhabi, UAE} \\
\email{umaima.rahman@mbzuai.ac.ae}
}

\begin{comment}  %% Removed for anonymized MICCAI 2025 submission
\author{First Author\inst{1}\orcidID{0000-1111-2222-3333} \and
Second Author\inst{2,3}\orcidID{1111-2222-3333-4444} \and
Third Author\inst{3}\orcidID{2222--3333-4444-5555}}
%
\authorrunning{F. Author et al.}
\institute{Princeton University, Princeton NJ 08544, USA \and
Springer Heidelberg, Tiergartenstr. 17, 69121 Heidelberg, Germany
\email{lncs@springer.com}\\
\url{http://www.springer.com/gp/computer-science/lncs} \and
ABC Institute, Rupert-Karls-University Heidelberg, Heidelberg, Germany\\
\email{\{abc,lncs\}@uni-heidelberg.de}}

\end{comment}

% \author{}  %% Added for anonymized MICCAI 2025 submission
\authorrunning{Rahman et al.}
    
\maketitle              % typeset the header of the contribution

\begin{abstract}  
Medical vision-language models (VLMs) offer promise for clinical decision support, yet their reliability under distribution shifts remains a major concern for safe deployment. These models often learn task-agnostic correlations due to variability in imaging protocols and free-text reports, limiting their generalizability and increasing the risk of failure in real-world settings.
We propose \textit{DRiFt}, a structured feature decoupling framework that explicitly separates clinically relevant signals from task-agnostic noise using parameter-efficient tuning (LoRA) and learnable prompt tokens. To enhance cross-modal alignment and reduce uncertainty, we curate high-quality, clinically grounded image-text pairs by generating captions for a diverse medical dataset.
Our approach improves in-distribution performance by +11.4\% Top-1 accuracy and +3.3\% Macro-F1 over prior prompt-based methods, while maintaining strong robustness across unseen datasets. Ablation studies reveal that disentangling task-relevant features and careful alignment significantly enhance model generalization and reduce unpredictable behavior under domain shift. These insights contribute toward building safer, more trustworthy VLMs for clinical use. The code is available at \href{https://github.com/rumaima/DRiFt}{https://github.com/rumaima/DRiFt}.

% \keywords{OOD Generalization, Vision-Language Models,  Decoupling Spurious Correlation, Parameter-Efficient FineTuning, Few shot Learning}
% Authors must provide keywords and are not allowed to remove this Keyword section.
\end{abstract}

\vspace{-0.5cm}

\begin{figure}[t]
\centering
\includegraphics[width=0.95\textwidth]{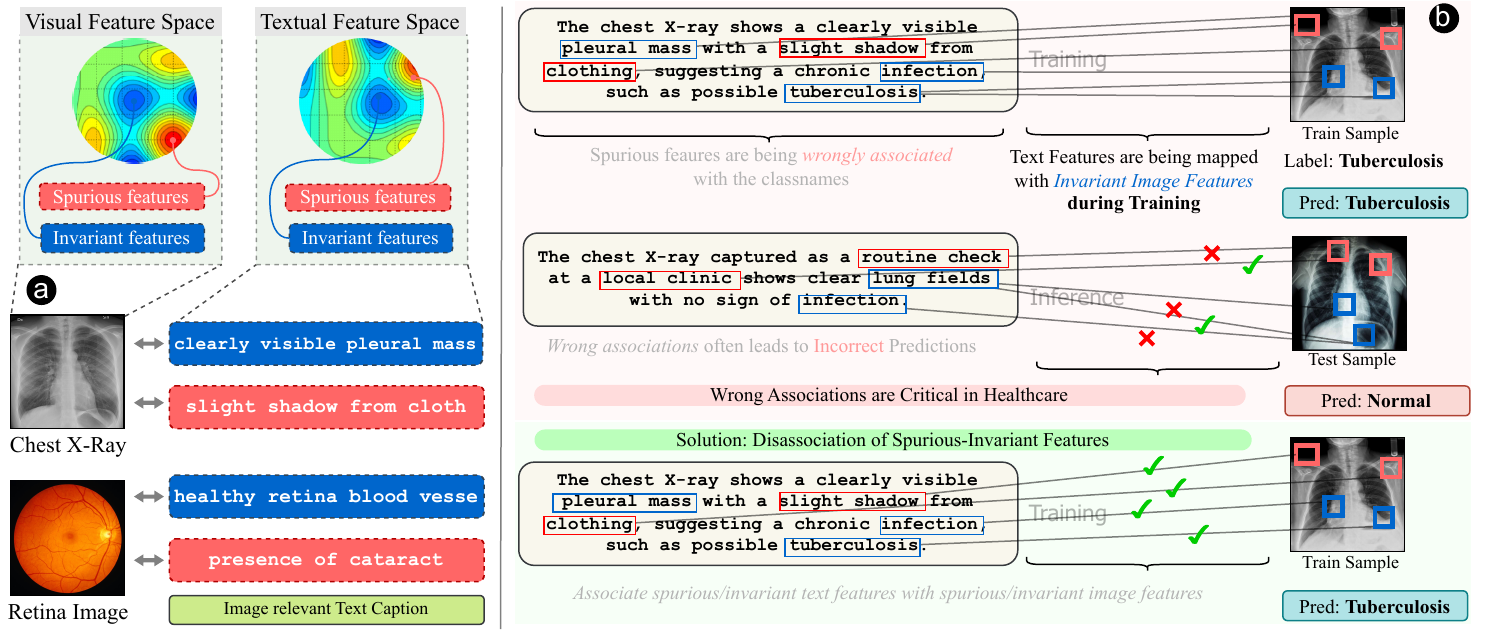}
\vspace{-0.2cm}
\caption{\small Impact of spurious correlations in medical imaging. \textbf{(a)} Conceptual illustration of multi-modal decoupling. \textbf{(b)} Shows how spurious features can dominate embeddings, leading to misclassification, critical in healthcare contexts.}
 \label{fig:concept}
\vspace{-0.5cm}
\end{figure}

\section{Introduction}

Recent developments in healthcare rely increasingly on automated medical imaging analysis to support critical diagnoses and treatment decisions \cite{sintchenko2003clinical}. However, real-world clinical scenarios often present substantial variability \cite{rahman2024mrishift} ranging from differences in patient demographics and imaging protocols to heterogeneous textual descriptions from radiology reports \cite{white2024heterogeneity}. These variations introduce \textit{spurious correlations} that can confound traditional vision-language models and degrade their performance when encountering domain shifts \cite{saab2022reducing}. Motivated by this, we propose a feature decoupling framework for vision-language models that targets both the visual and textual domains, explicitly separating clinically relevant, i.e., \textit{invariant} features, from spurious artifacts. As illustrated in Fig. \ref{fig:concept}, non-clinical details such as the presence of clothing shadows or incidental notes about patient scheduling, can mislead a model if these spurious elements become incorrectly associated with diagnostic findings (e.g., lung infiltrates) \cite{compton2023more}.

Conventional fine-tuning strategies such as \cite{khattak2023maple,zhou2022conditional} that treat vision and text holistically risk conflating essential signals with irrelevant context, undermining spurious relationships which can be very sensitive in high-stakes medical settings. Building on \cite{rahman2025can}, who align unpaired images and texts for domain robustness, we leverage caption-driven supervision to enhance clinical relevance while maintaining feature disentanglement. Inspired by the success of our prior work on disentangled prompt learning in natural images \cite{rahman2025dimple} and its relevance to medical imaging, we extend the approach to address clinical domain shift. We aim to enforce a structured decoupling of features in both modalities. Crucially, spurious image features must map to spurious text descriptors, while clinically pertinent cues in the image remain aligned with relevant textual annotations. Through parameter-efficient fine-tuning (via LoRA \cite{hu2022lora,imam2024test}) and systematic image-caption curation within MedIMeta \cite{woerner2024comprehensive}, we show that this \textit{decoupled approach} not only improves generalization under cross-dataset and domain generalization conditions but also reduces the computational burden of adapting large multi-modal models \cite{dai2023instructblipgeneralpurposevisionlanguagemodels}.
% By prioritizing invariant feature alignment and isolating spurious signals, our framework delivers more trustworthy and resilient AI-driven diagnostics for real-world healthcare environments. 
Our contributions can be summarized as follows:
\vspace{-2mm}
\begin{enumerate}
    \item \textbf{Structured Feature Decoupling Across Modalities:} We design a framework that separates image and text embeddings into invariant and spurious components, ensuring that clinically relevant signals remain unaffected by domain-specific artifacts. This minimizes erroneous cross-modal associations and enhances generalization in healthcare settings.
    \item \textbf{Parameter-Efficient LoRA Fine-Tuning:} To address computational constraints, we employ LoRA-based fine-tuning, updating only low-rank adapter layers instead of the entire model. By optimizing contrastive and cross-modal alignment losses, our framework mitigates domain biases and improves robustness to out-of-distribution data.
    \item \textbf{Enriching MedIMeta with Caption Generation for Few-Shot Training:} We leverage the MedIMeta dataset covering multiple imaging modalities to generate clinically relevant captions. This ensures the model learns invariant medical patterns without being misled by domain-specific artifacts.
\end{enumerate}

% In this work, we propose a new decoupled multi-modal learning framework tailored to address the above challenges. By (1) explicitly separating invariant, clinically relevant features from dataset-specific artifacts and (2) integrating prompt-based adaptation for both image and text representations, our approach aims to deliver robust performance under domain shifts. This method stands out by combining domain generalization and multi-modal learning principles, highlighting its potential to advance AI-driven healthcare solutions that remain trustworthy and effective across diverse imaging datasets and clinical scenarios.

\noindent\textbf{Related Works:}
As healthcare data diversifies, domain generalization strategies \cite{li2022domain,yoon2024domain,guo2022evaluation} become essential for robust performance across patient populations and imaging conditions.
Modern vision-language models (VLMs) enable few-shot or zero-shot inference by jointly encoding text and images but risk carrying over spurious correlations \cite{shakeri2024few,lai2023clipath,imam2025noise}. While prompt engineering provides a parameter-efficient adaptation strategy \cite{rahman2024meduna}, it rarely disentangles clinically relevant features from domain-specific noise, leading to overfitting \cite{liu2022learning}.
Recent unified multi-modal frameworks \cite{khattak2023maple} enhance image-text alignment but seldom separate spurious correlations from invariant clinical markers across both modalities, limiting reliability in real-world diagnostics. In contrast, our work emphasizes \textit{explicit feature decoupling} to reduce reliance on non-clinical details, ensuring more robust and interpretable medical learning algorithms.

% \begin{figure}[t]
% \includegraphics[width=\textwidth]{fig1.eps}
% \caption{A figure caption is always placed below the illustration.
% Please note that short captions are centered, while long ones are
% justified by the macro package automatically.} \label{fig:concept}
% \end{figure}

\section{Methodology}
\begin{figure}[t]
\includegraphics[width=\textwidth]{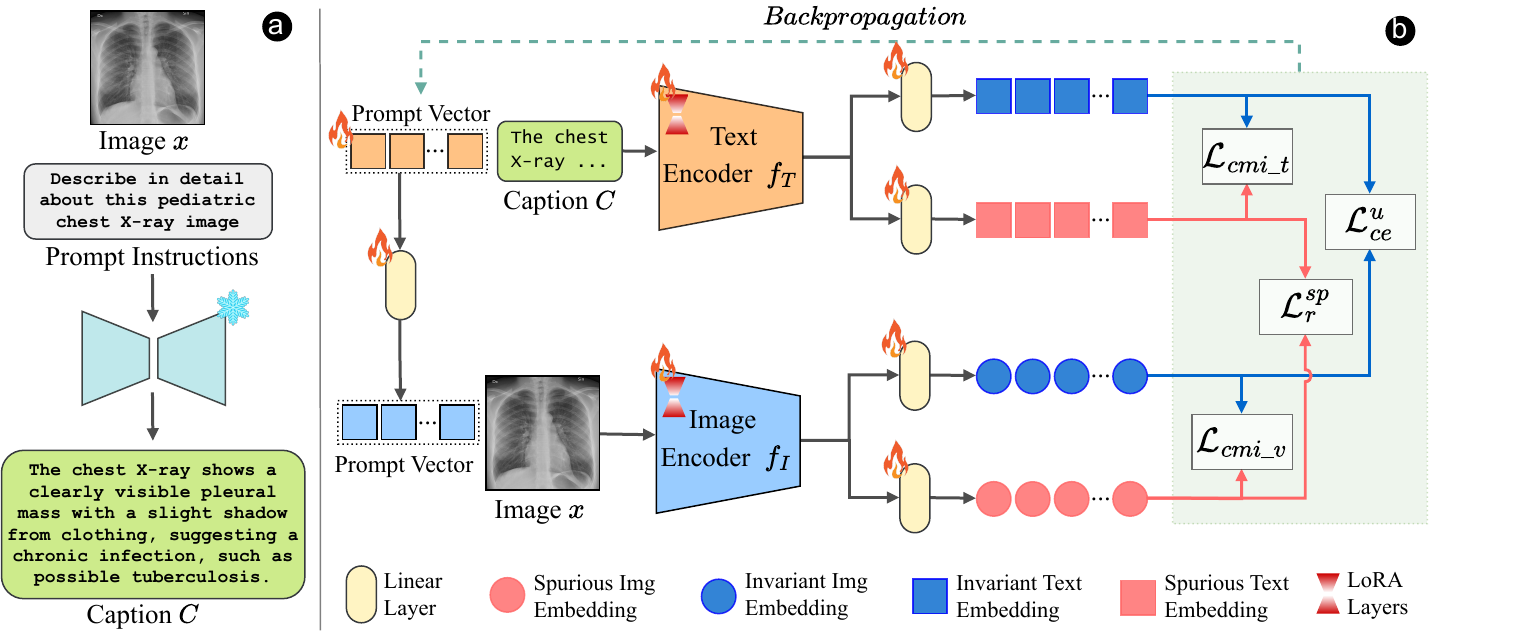}
% \captionsetup{font=scriptsize}
\caption{\small DRiFt: Our proposed feature disentanglement framework. \textbf{(a)} Captions are generated for images from the MedIMeta dataset. \textbf{(b)} Multi-modal feature decoupling is applied using LoRA fine-tuning and learnable prompts to focus on invariant features.}
 \label{fig:med_dimple_concept}
\vspace{-0.5cm}
\end{figure}

Medical imaging data in \textit{MedIMeta} \cite{woerner2024comprehensive} spans multiple modalities (e.g., chest X-ray, fundus, ultrasound), each subject to domain-specific artifacts and protocol variations. We introduce a decoupled multi-modal framework that separates invariant, clinically relevant features from spurious correlations in both vision and text modalities. This is achieved through caption generation and parameter-efficient adaptation using LoRA-based fine-tuning in a few-shot setting as illustrated in Fig. \ref{fig:med_dimple_concept}

\vspace{-0.2cm}
\subsection{Workflow}
\textbf{Caption Generation for MedIMeta:}
We generate captions for selected datasets within MedIMeta, such as skin-lesion and fundus images, ensuring a minimum image count threshold (e.g., 100+) for diversity. Captions are generated using InstructBLIP \cite{dai2023instructblipgeneralpurposevisionlanguagemodels}, prioritizing clinical relevance while preserving minor spurious details. Among these captions, we validated their clinical relevance, removing those that were irrelevant while retaining the meaningful contextual elements. A subset of these captions is selected for a few-shot setting, aiding the model in distinguishing clinically meaningful information from spurious correlations.

\textbf{Decoupled Feature Representation:}
To enhance generalization, we decompose each image and text embedding into invariant and spurious components \cite{rahman2025dimple}. Given an input embedding $\mathbf{z_M}_i$, two projection functions are defined by \(\mathbf{z_M}_{i,u} = \phi_M(\mathbf{z_M}_i)\), 
where $M={\mathbf{v}, \mathbf{t}}$, with $\mathbf{v}$ and $\mathbf{t}$ representing visual and textual embeddings, respectively. This decomposition ensures clinically meaningful features are preserved while mitigating domain-specific artifacts.
% \cite{sarafraz2024domain,yoon2024domain}.

\textbf{Multi-Modal Low-Rank Adaptation for Image-Caption Learning:}
We introduce \textit{LoRA} adapters \cite{hu2022lora} to efficiently update the query ($Q$) and key ($K$) projections of the vision and text encoders. Instead of fine-tuning the entire Transformer layers, LoRA injects low-rank matrices into the original weight matrix $W_0 \in \mathbb{R}^{d_{\text{out}} \times d_{\text{in}}}$, decomposing it as, \(
    W_{\theta} = W_0 + \alpha A B, \)
where $A \in \mathbb{R}^{d_{\text{out}} \times r}$, $B \in \mathbb{R}^{r \times d_{\text{in}}}$, and $\theta = \{A, B\}$ denotes the trainable LoRA parameters. Alongside LoRA, we integrate \textit{learnable prompt tokens} $\gamma = \{p_1, p_2, ..., p_m\}$, where $P_{\gamma} \in \mathbb{R}^{m \times d}$, inserted at the input layer to guide the model towards clinically meaningful patterns while suppressing spurious correlations \cite{sarafraz2024domain}. The overall trainable parameter set is \(
    \Theta = \theta \cup \gamma.
\)

\vspace{-0.2cm}
\subsection{Objectives}
To ensure the model captures clinically relevant features, we define an objective function comprising:

\noindent\textbf{1. Invariant Alignment}: A contrastive loss $\mathcal{L}^{u}_{ce}$ aligns visual and textual invariant embeddings:
\vspace{-2mm}
\begin{equation}
\small
   \mathcal{L}^{u}_{ce} = 
    - \sum_{i=1}^{N} y_i \log p^u_\Theta(y_i \mid x_i), \quad
    p^u_\Theta(y_i \mid x_i) = 
    \frac{\exp(\texttt{sim}(\mathbf{z_v}_{i,u}, \mathbf{z_t}_{i,u}) / \tau)}{\sum_{j=1}^{C} \exp(\texttt{sim}(\mathbf{z_v}_{j,u}, \mathbf{z_t}_{j,u}) / \tau)}. \nonumber
\end{equation}

\noindent\textbf{2. Spurious Neutralization}: A KL-divergence loss $\mathcal{L}^{sp}_r$ prevents spurious features across both modalities from influencing classification:
\vspace{-2mm}
\begin{equation}
\small
    \mathcal{L}^{sp}_r = 
    \sum_{i=1}^{N} \ell_{KL}(p^s_\Theta(y_i \mid x_i) \mid\mid p_0).
\end{equation}

\noindent\textbf{3. Reduced Statistical Dependence}: We enforce conditional independence between the invariant and textual features across vision and textual modalities using the Conditional Hilbert-Schmidt Independence Criterion:
\vspace{-1mm}

\begin{equation}
\small
\mathcal{L}_{con_v} = I(\mathbf{z_v}_{i,u}; \mathbf{z_v}_{i,s} \mid Y), 
\quad 
\mathcal{L}_{con_t} = I(\mathbf{z_t}_{i,u}; \mathbf{z_t}_{i,s} \mid Y).
\end{equation}
% The final independence loss is computed as:
% % \vspace{-2mm}
% \begin{equation}
% \mathcal{L}_{con} = \text{avg}(\mathcal{L}_{con_v}, \mathcal{L}_{con_t}).
% \end{equation}
The overall loss function integrates these components:
% \vspace{-2mm}
\begin{equation}
\small
    \mathcal{L} = \mathcal{L}^{u}_{ce} + \alpha \mathcal{L}^{sp}_r + \beta \mathcal{L}_{con}, ~~\text{where}~~ \mathcal{L}_{con} = \text{avg}(\mathcal{L}_{con_v}, \mathcal{L}_{con_t}).
\end{equation}
where $\alpha$ and $\beta$ regulate spurious suppression and statistical independence respectively while $\mathcal{L}_{con}$ is the independence loss. By leveraging curated captions, decoupled embeddings, LoRA-enhanced prompts, and the above objective functions, DRiFt achieves comparatively better generalization under real-world distribution shifts in medical imaging as seen in Figure \ref{fig:ablation_1}(left).

\vspace{-4mm}

\section{Experimentation and Results}
\subsection{Experimental Setup}
\vspace{-1.4mm}
\textbf{Datasets:} We assess DRiFt on 8 different MedIMeta datasets \cite{woerner2024comprehensive} belonging to 6 different medical modalities, such as chest-Xray, breast ultrasound, mammography, dermatoscopy, and fundus images. We also evaluate our models in a cross-dataset generalization setting for the datasets \texttt{S-TB} \cite{shenzhenTB}, \texttt{M-TB} \cite{montgomeryTB}, \texttt{IDRID} \cite{porwal2018indian}, and \texttt{ISIC} \cite{isicArchive}. We choose \textit{MedIMeta} dataset as it encompasses multi-classification tasks which provides a rigorous test bed for evaluating model resilience to distribution shifts to ensure better cross-domain performance in real-world clinical scenarios \cite{imam2025robustness}.

{\renewcommand{\arraystretch}{1.0}
\begin{table}[t]
\caption{Comparison of Top-1 Accuracy,  Macro-F1, and AUC on in-distribution data using our methodology - DRiFt with respect to other multi-modal prompt learning approaches adapted to an image-caption pair setting for fair comparison.}\label{main_table}
\centering
\resizebox{\textwidth}{!}{ 
\begin{tabular}{c|l|ccc|ccc|ccc}
\toprule
\# Classes & Methods $\rightarrow$ & \multicolumn{3}{c}{\texttt{MaPLe-IC}} & \multicolumn{3}{|c}{\texttt{CoOp-OOD-IC}} & \multicolumn{3}{|c}{\texttt{DRiFt}}\\ 
\cmidrule(r){2-2} \cmidrule(lr){3-5} \cmidrule(lr){6-8} \cmidrule(l){9-11} 
& MedIMeta Dataset \cite{woerner2024comprehensive} & \cellcolor[HTML]{f8f8fd}Acc. & \cellcolor[HTML]{f8f8fd}Macro-F1 & \cellcolor[HTML]{f8f8fd}AUC &  \cellcolor[HTML]{f8f8fd}Acc. & \cellcolor[HTML]{f8f8fd}Macro-F1  & \cellcolor[HTML]{f3effc}AUC & \cellcolor[HTML]{f3effc}Acc. & \cellcolor[HTML]{f3effc}Macro-F1  & \cellcolor[HTML]{f3effc}AUC \\ 
\midrule
3 & \texttt{bus}            &\cellcolor[HTML]{f8f8fd}11.5  &\cellcolor[HTML]{f8f8fd}10.1 &\cellcolor[HTML]{f8f8fd}29.3 &\cellcolor[HTML]{f8f8fd}17.3    & \cellcolor[HTML]{f8f8fd}9.8   &\cellcolor[HTML]{f8f8fd}\textbf{50.5} &   \cellcolor[HTML]{f3effc}\textbf{42.3} & \cellcolor[HTML]{f3effc}\textbf{21.7}   &\cellcolor[HTML]{f3effc}34.9  \\
15 & \texttt{skinl-derm}     &\cellcolor[HTML]{f8f8fd}1.3   &\cellcolor[HTML]{f8f8fd}0.2  &\cellcolor[HTML]{f8f8fd}\textbf{52.1} &\cellcolor[HTML]{f8f8fd}4.1     & \cellcolor[HTML]{f8f8fd}2.1   &\cellcolor[HTML]{f8f8fd}48.0 &   \cellcolor[HTML]{f3effc}\textbf{21.5} & \cellcolor[HTML]{f3effc}\textbf{3.2 }   &\cellcolor[HTML]{f3effc}49.5  \\
7 & \texttt{derm}           &\cellcolor[HTML]{f8f8fd}34.1  &\cellcolor[HTML]{f8f8fd}10.1 &\cellcolor[HTML]{f8f8fd}\textbf{46.8} &\cellcolor[HTML]{f8f8fd}16.9    & \cellcolor[HTML]{f8f8fd}6.1   &\cellcolor[HTML]{f8f8fd}45.6 &   \cellcolor[HTML]{f3effc}\textbf{57.3} & \cellcolor[HTML]{f3effc}\textbf{11.6}   &\cellcolor[HTML]{f3effc}46.5  \\
2 & \texttt{glaucoma}       &\cellcolor[HTML]{f8f8fd}84.8  &\cellcolor[HTML]{f8f8fd}45.9 &\cellcolor[HTML]{f8f8fd}56.3 &\cellcolor[HTML]{f8f8fd}15.2    & \cellcolor[HTML]{f8f8fd}13.2  &\cellcolor[HTML]{f8f8fd}\textbf{60.7} &   \cellcolor[HTML]{f3effc}\textbf{82.1} &  \cellcolor[HTML]{f3effc}\textbf{52.2}  &\cellcolor[HTML]{f3effc}51.2  \\
2 & \texttt{fundus}         &\cellcolor[HTML]{f8f8fd}21.1  &\cellcolor[HTML]{f8f8fd}17.5 &\cellcolor[HTML]{f8f8fd}\textbf{68.7} &\cellcolor[HTML]{f8f8fd}20.9    & \cellcolor[HTML]{f8f8fd}17.3  &\cellcolor[HTML]{f8f8fd}21.1 &   \cellcolor[HTML]{f3effc}\textbf{22.2} &  \cellcolor[HTML]{f3effc}\textbf{19.0}  &\cellcolor[HTML]{f3effc}59.1  \\
3 & \texttt{pneumonia}      &\cellcolor[HTML]{f8f8fd}30.3  &\cellcolor[HTML]{f8f8fd}26.3 &\cellcolor[HTML]{f8f8fd}47.6 &\cellcolor[HTML]{f8f8fd}37.5    & \cellcolor[HTML]{f8f8fd}18.2  &\cellcolor[HTML]{f8f8fd}28.4 &   \cellcolor[HTML]{f3effc}\textbf{37.5} &  \cellcolor[HTML]{f3effc}18.2           &\cellcolor[HTML]{f3effc}\textbf{52.4}  \\
2 & \texttt{mammo\_mass}    &\cellcolor[HTML]{f8f8fd}\textbf{50.5}  &\cellcolor[HTML]{f8f8fd}\textbf{49.1} &\cellcolor[HTML]{f8f8fd}\textbf{46.5} &\cellcolor[HTML]{f8f8fd}38.6    & \cellcolor[HTML]{f8f8fd}27.9  &\cellcolor[HTML]{f8f8fd}42.1 &   \cellcolor[HTML]{f3effc}50.3          &  \cellcolor[HTML]{f3effc}48.3           &\cellcolor[HTML]{f3effc}45.9  \\
2 & \texttt{mammo\_calc}    &\cellcolor[HTML]{f8f8fd}41.4  &\cellcolor[HTML]{f8f8fd}40.8 &\cellcolor[HTML]{f8f8fd}48.0 &\cellcolor[HTML]{f8f8fd}41.1    & \cellcolor[HTML]{f8f8fd}32.9  &\cellcolor[HTML]{f8f8fd}\textbf{60.4} &   \cellcolor[HTML]{f3effc}\textbf{52.8} &  \cellcolor[HTML]{f3effc}\textbf{52.0}  &\cellcolor[HTML]{f3effc}53.3  \\
\midrule
 & \texttt{average}        &\cellcolor[HTML]{f8f8fd}34.4  &\cellcolor[HTML]{f8f8fd}25.0 &\cellcolor[HTML]{f8f8fd}\textbf{49.4} &\cellcolor[HTML]{f8f8fd}24.0    & \cellcolor[HTML]{f8f8fd}15.9  &\cellcolor[HTML]{f8f8fd}44.6 &   \cellcolor[HTML]{f3effc}\textbf{45.8} & \cellcolor[HTML]{f3effc}\textbf{28.3}   &\cellcolor[HTML]{f3effc}49.1  \\
\bottomrule
\end{tabular}}
\end{table}

{\renewcommand{\arraystretch}{1.0}
\begin{table}[t]
 \vspace{-5mm}
\caption{Cross-dataset evaluation comparing MaPLe-IC, CoOp-OOD-IC, DRiFt. }
\label{cross_table}
\centering
\resizebox{\textwidth}{!}{ 
\begin{tabular}{l|cc|cc|cc}
\toprule
{\textbf{Method}}$\rightarrow$ & \multicolumn{2}{c|}{\textbf{MaPLe-IC}} & \multicolumn{2}{c|}{\textbf{CoOp-OOD-IC}} & \multicolumn{2}{c}{\textbf{DRiFt}} \\
\cmidrule(lr){1-3} \cmidrule(lr){4-5} \cmidrule(l){6-7}  
MedIMeta Dataset & Target & Accuracy & Target & Accuracy & Target & Accuracy \\
\midrule
\texttt{skinl\_derm}  & \cellcolor[HTML]{f8f8fd}avg(derm, isic) & \cellcolor[HTML]{f8f8fd}6.4  & \cellcolor[HTML]{f8f8fd}avg(derm, isic) & \cellcolor[HTML]{f8f8fd}6.5  & \cellcolor[HTML]{f3effc}avg(derm, isic) & \cellcolor[HTML]{f3effc}10.4  \\
\midrule
\texttt{derm}       & \cellcolor[HTML]{f8f8fd}avg(skinl\_derm, isic) & \cellcolor[HTML]{f8f8fd}18.9  & \cellcolor[HTML]{f8f8fd}avg(skinl\_derm, isic) & \cellcolor[HTML]{f8f8fd}18.8  & \cellcolor[HTML]{f3effc}avg(skinl\_derm, isic) & \cellcolor[HTML]{f3effc}17.8  \\
\midrule
\texttt{glaucoma}      & \cellcolor[HTML]{f8f8fd}avg(fundus, idrid) & \cellcolor[HTML]{f8f8fd}24.9  & \cellcolor[HTML]{f8f8fd}avg(fundus, idrid) & \cellcolor[HTML]{f8f8fd}25.7  & \cellcolor[HTML]{f3effc}avg(fundus, idrid) & \cellcolor[HTML]{f3effc}24.8  \\
\midrule
\texttt{fundus}       & \cellcolor[HTML]{f8f8fd}avg(glaucoma, idrid) & \cellcolor[HTML]{f8f8fd}55.6  & \cellcolor[HTML]{f8f8fd}avg(glaucoma, idrid) & \cellcolor[HTML]{f8f8fd}55.6  & \cellcolor[HTML]{f3effc}avg(glaucoma, idrid) & \cellcolor[HTML]{f3effc}56.4  \\
\midrule
\texttt{pneumonia}    & \cellcolor[HTML]{f8f8fd}avg(s-tb, m-tb) & \cellcolor[HTML]{f8f8fd}54.3  & \cellcolor[HTML]{f8f8fd}avg(s-tb, m-tb) & \cellcolor[HTML]{f8f8fd}51.8  & \cellcolor[HTML]{f3effc}avg(s-tb, m-tb) & \cellcolor[HTML]{f3effc}52.2  \\
\bottomrule
\end{tabular}} 
\vspace{-0.5cm}
\end{table}

\textbf{Baselines: }  
We compare DRiFt against two other multi-modal foundation methods aimed at out-of-distribution (OOD) generalization in the vision domain. In particular, we adapt \textit{MaPLe} \cite{khattak2023maple} and \textit{CoOp-OOD} \cite{zhangamend} to accommodate image-caption pairs and integrate LoRA-based fine-tuning for parameter efficiency. MaPLe originally improved generalization by conditioning visual prompts on textual prompts, while CoOp-OOD enhanced robustness by decoupling visual features and aligning invariant representations with text. 
However, these methods do not explicitly separate spurious features within their textual counterparts. To address this, we introduce a unified decoupling framework that ensures clinically meaningful features are preserved while mitigating domain-specific artifacts.

\textbf{Implementation Details: }  DRiFt adopts a few-shot learning setup by selecting 16 random samples per class for training. A pre-trained ViT-B/16-based vision-language model serves as the backbone, using dimension settings of \(\textit{d}_l = 512\) for language, \(\textit{d}_v = 768\) for vision, and \(\textit{d}_{vl} = 512\) for combined embeddings. In total, we employ $J=3$ transformer layers of prompt tuning, each with two learnable prompt tokens for vision and language. We train on a single NVIDIA A100 GPU for 10 epochs using stochastic gradient descent with a batch size of 4 and an initial learning rate of 2.6e-6. For initialization, the first-layer text prompt vectors are taken from a pre-trained word embedding (a photo of a <class>) and concatenated with the caption corresponding to the image, while subsequent layers are randomly initialized from a normal distribution.

\textbf{Evaluation Metrics: }  Table \ref{main_table} presents the evaluation of different medical datasets using \textit{Accuracy}, \textit{Macro F1-score}, and \textit{ROC-AUC score}. Accuracy is the percentage of correctly classified instances and provides an overall performance measure, howvere, it may be misleading for imbalanced datasets. Macro F1-score is the average of F1-scores across all classes. It treats all classes equally, making it more informative when class distributions are skewed. Furthermore, a higher ROC-AUC score indicates better classification performance and measures the model's ability to distinguish between classes by plotting the true positive rate (TPR) against the false positive rate (FPR) at different thresholds. \\

% Performance is reported along with their harmonic mean, to capture a balanced view of generalization under domain shifts.
\begin{figure*}[t!]
    \centering
    \begin{minipage}[t]{0.22\linewidth}
        \centering
        \includegraphics[width=\linewidth]{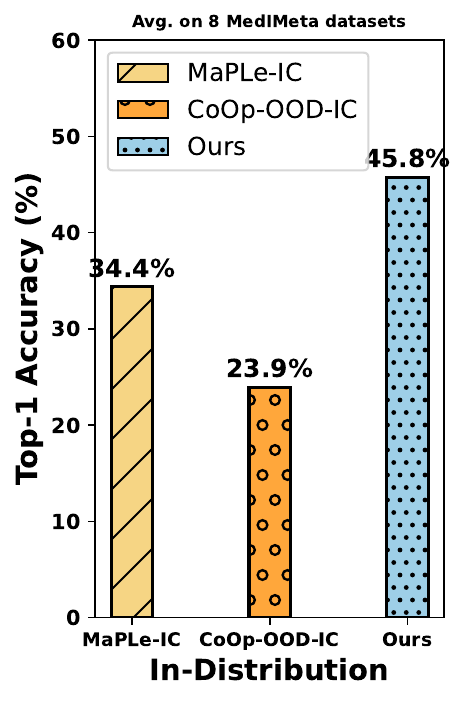}
        \label{fig:performance_ind}
    \end{minipage}\hfill
    \begin{minipage}[t]{0.37\linewidth}
        \centering
        \includegraphics[width=\linewidth]{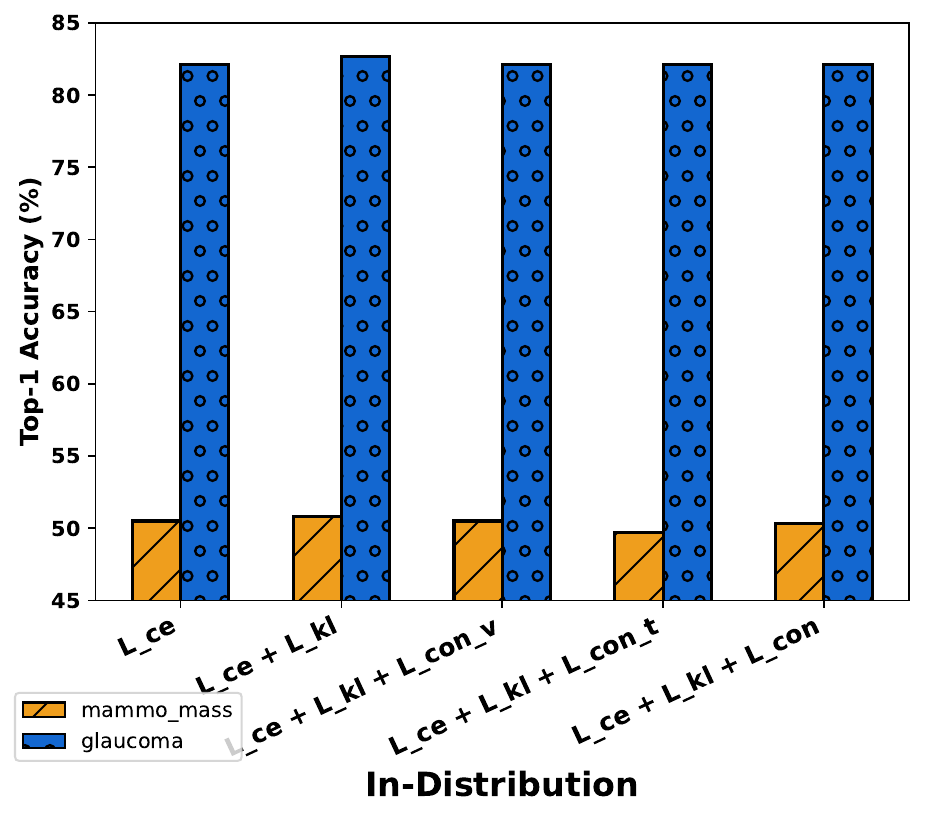}
        \label{fig:loss_ind}
    \end{minipage}\hfill
    \begin{minipage}[t]{0.38\linewidth}
        \centering
        \includegraphics[width=\linewidth]{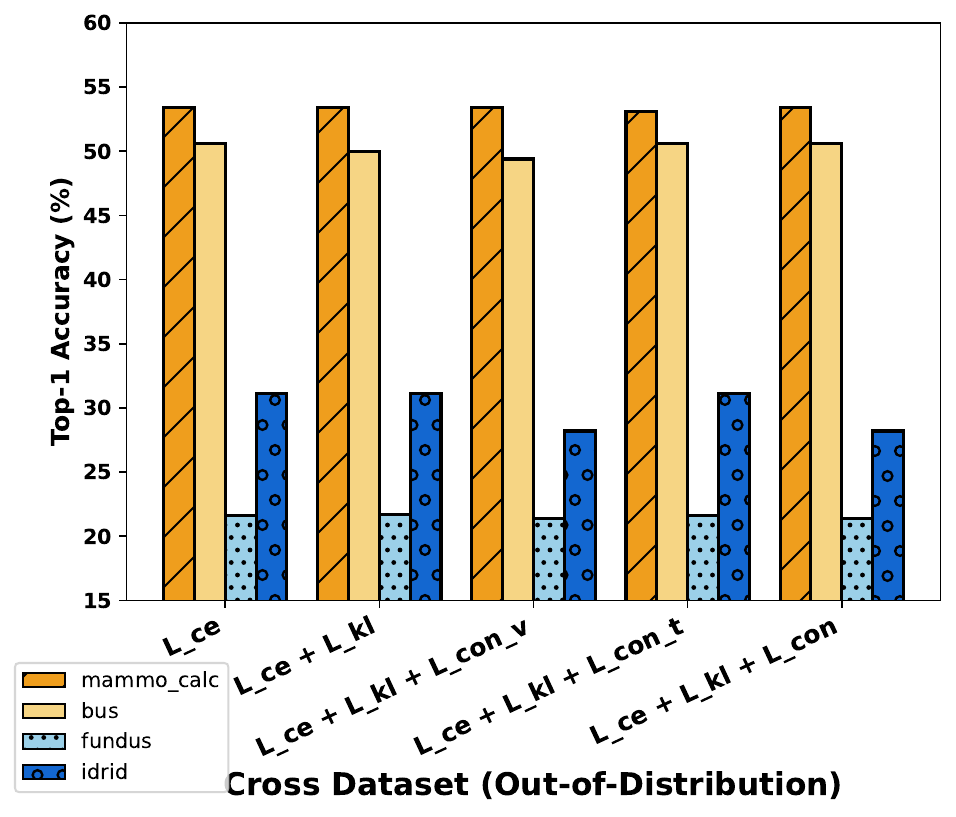}
        \label{fig:loss_ood}
    \end{minipage}\hfill
    \vspace{-5mm}
    \label{fig:in_distribution}
    % \captionsetup{font=scriptsize}
    \caption {\small \textbf{(left)} In-distribution performance comparison across MaPLe-IC, CoOp-OOD-IC, and DRiFt, with ours achieving the highest average Top-1 accuracy across 8 MedIMeta tasks. \textbf{(middle)} Impact of loss components on in-distribution performance, showing Top-1 accuracy variations for the mammo\_mass and glaucoma tasks. \textbf{(right)} Cross-dataset evaluation of loss components, highlighting Top-1 accuracy changes.}
    \vspace{-7mm}
\end{figure*}

\vspace{-10mm}
\subsection{Main Results}
\textbf{In-Distribution Evaluation:} 
As shown in Table \ref{main_table}, DRiFt outperforms MaPLe and CoOpOOD across multiple datasets, achieving the highest accuracy in bus (42.3\%), skinl\_derm (21.5\%), and mammo\_calc (52.8\%). It also achieves the best Macro-F1 in glaucoma (52.2\%), mammo\_calc (52\%) and bus (21.7\%), ensuring better class balance. While pneumonia accuracy improves (37.5\%), its low Macro-F1 (18.2\%) highlights challenges in minority class predictions. On average DRiFt surpasses MaPLe and CoOp-OOD with 45.75\% accuracy and 28.275 Macro-F1, demonstrating superior generalization in medical classification.

\textbf{Cross-Dataset Evaluation:}  
The cross-dataset setting assesses model generalization by training on one dataset and testing on another with similar modalities but distinct characteristics. This evaluates how well the model adapts to variations in imaging protocols and anatomical structures. For skin lesion classification, DRiFt outperforms MaPLe and CoOp-OOD on skinl\_derm (10.35\% vs. 6.5\% and 6.45\%) but slightly underperforms on derm (17.75\% vs. 18.95\%) as tabulated in Table \ref{cross_table}.

\vspace{-5mm}
\section{Ablation Study}
\vspace{-5mm}
\textbf{a. Effect of Number of shots:}
Increasing the number of shots generally improves performance (Fig. \ref{fig:ablation_2}(left)), particularly in skinl\_derm (7.8\% to 21.5\%), derm (57.3\% stable), and bus (50\% to 42.3\%), suggesting better class separability with more data. Glaucoma (83.9\% to 82.1\%) remains mostly unchanged, while pneumonia (37.5\%) is entirely unaffected, indicating that some tasks do not benefit from additional samples. This suggest that while increased shots enhance fine-grained classification, its efficacy depends on dataset complexity and class distribution.
\begin{figure*}[t!]
 % % \vspace{-7mm}
    \centering
    \begin{minipage}[t]{0.52\linewidth}
        \centering
        \includegraphics[width=0.9\linewidth]{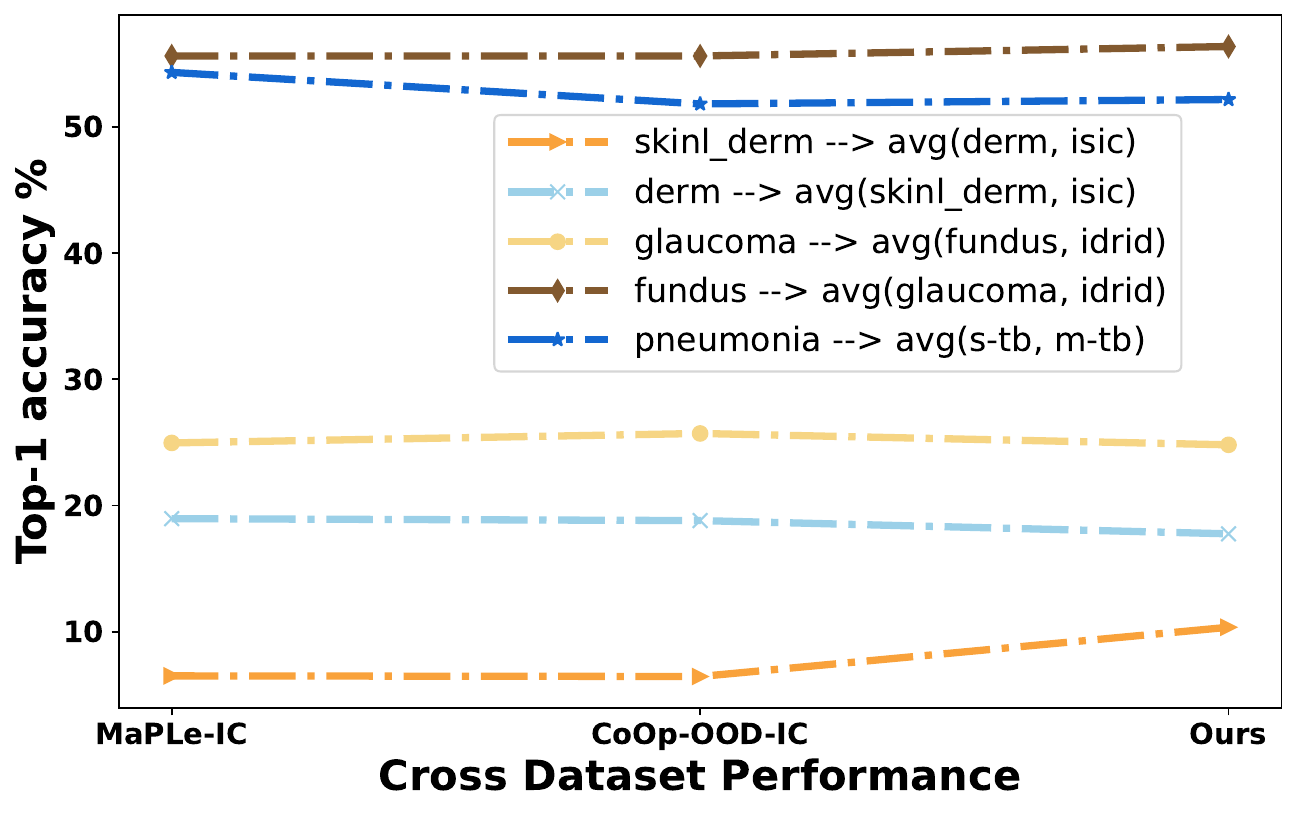}
        \label{fig:cross_dataset}
    \end{minipage}\hfill
    \begin{minipage}[t]{0.23\linewidth}
        \centering
        \includegraphics[width=\linewidth]{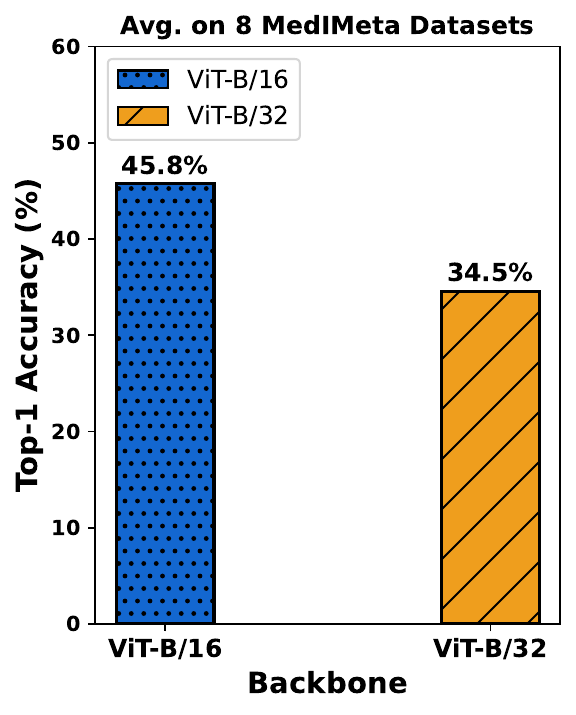}
        \label{fig:backbone}
    \end{minipage}\hfill
    \begin{minipage}[t]{0.23\linewidth}
        \centering
        \includegraphics[width=\linewidth]{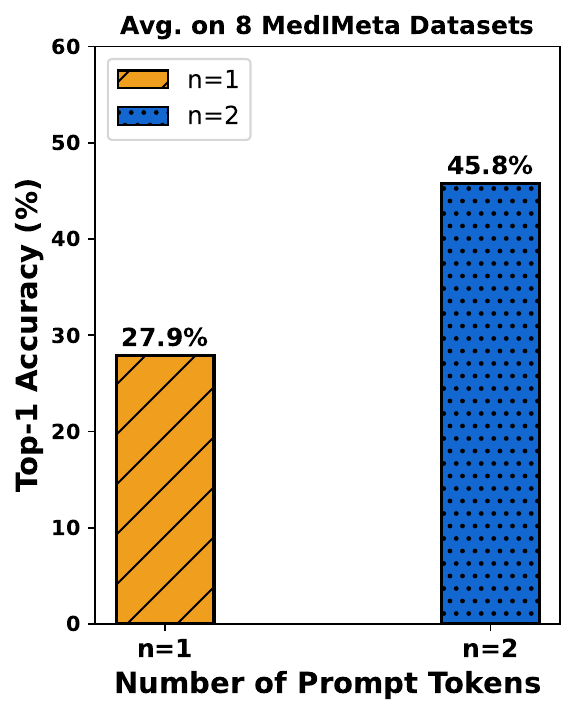}
        \label{fig:tokens}
    \end{minipage}\hfill
    \vspace{-5mm}
    % \captionsetup{font=scriptsize}
    \caption {\small \textbf{(left)} Comparison of cross-dataset generalization for MaPLe-IC, CoOp-OOD-IC, and DRiFt (Ours). \textbf{(middle)} Comparison of ViT-B/16 and ViT-B/32 backbones on MedIMeta datasets, showing that ViT-B/16 achieves higher average Top-1 accuracy. \textbf{(right)} Increasing the number of prompt tokens from 1 to 2 significantly improves the average Top-1 accuracy across MedIMeta datasets, from 27.9\% to 45.8\%.}
    \label{fig:ablation_1}
\vspace{-5mm}
\end{figure*}

\noindent\textbf{b. Effect of Number of prompt tokens: }
Using two prompt tokens ($n=2$) consistently improves performance over a single token ($n=1$), particularly in bus (17.3\% to 42.3\%), skinl\_derm (8.1\% to 21.5\%), and derm (30.9\% to 57.3\%), suggesting that additional tokens enrich feature extraction. Similarly, mammo\_mass (38.6\% to 50.3\%) and mammo\_calc (41.1\% to 52.8\%) benefit from more tokens, aiding complex imaging tasks. However, fundus (33.6\% to 22.2\%) experiences a decline, likely due to redundant or non-discriminative information, while pneumonia (37.5\%) remains unaffected. On average, increasing tokens enhances generalization, but tuning is essential based on the dataset.

\noindent\textbf{c. Effect of Number of prompt depth: }
Increasing prompt depth generally degrades performance, with bus (42.3\% to 16\%), skinl\_derm (21.5\% to 1.5\%), and derm (57.3\% to 21.8\%) showing steep declines. Glaucoma (82.1\% to 24.7\%) initially drops but partially recovers at $J=12$ (46.1\%), indicating instability. While some datasets, like fundus (22.2\% to 25.9\%), show marginal gains, deeper prompts tend to introduce redundancy, making careful selection crucial.

\noindent\textbf{d. Effect of backbone:}
ViT-B/16 consistently outperforms ViT-B/32 in bus (42.3\% vs. 9\%), skinl\_derm (21.5\% vs. 4.8\%), and derm (57.3\% vs. 12.3\%), demonstrating the advantage of higher spatial resolution in fine-grained tasks. However, glaucoma (82.1\% vs. 85.1\%) remains unaffected, and mammo\_mass (50.3\% vs. 60.8\%) and mammo\_calc (52.8\% vs. 59.8\%) favor ViT-B/32, suggesting that backbone choice should depend on task requirements, balancing local detail extraction and global contextual representation.

\begin{figure*}[t!]
% \vspace{-4mm}
    \centering
    \begin{minipage}[t]{0.47\linewidth}
        \centering
        \includegraphics[width=\linewidth]{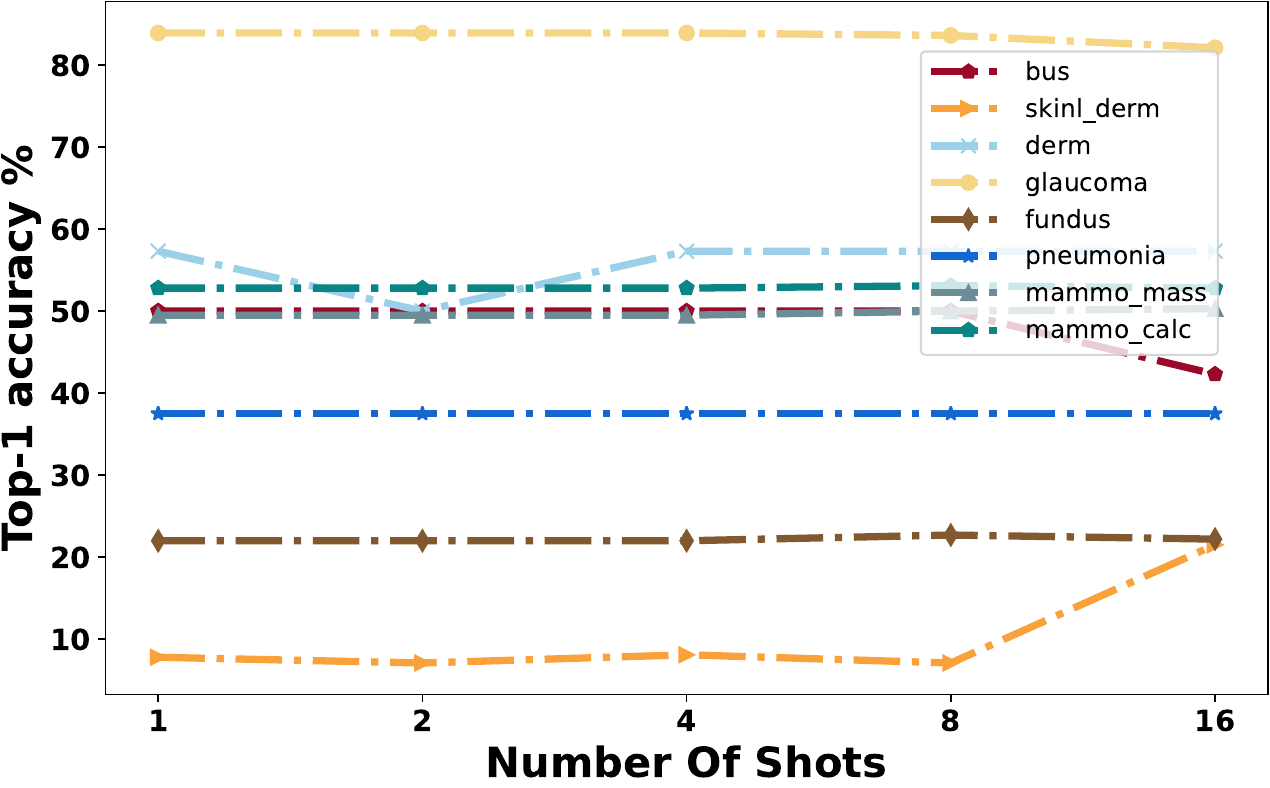}
        \label{fig:num_shots}
    \end{minipage}\hfill
    \begin{minipage}[t]{0.47\linewidth}
        \centering
        \includegraphics[width=\linewidth]{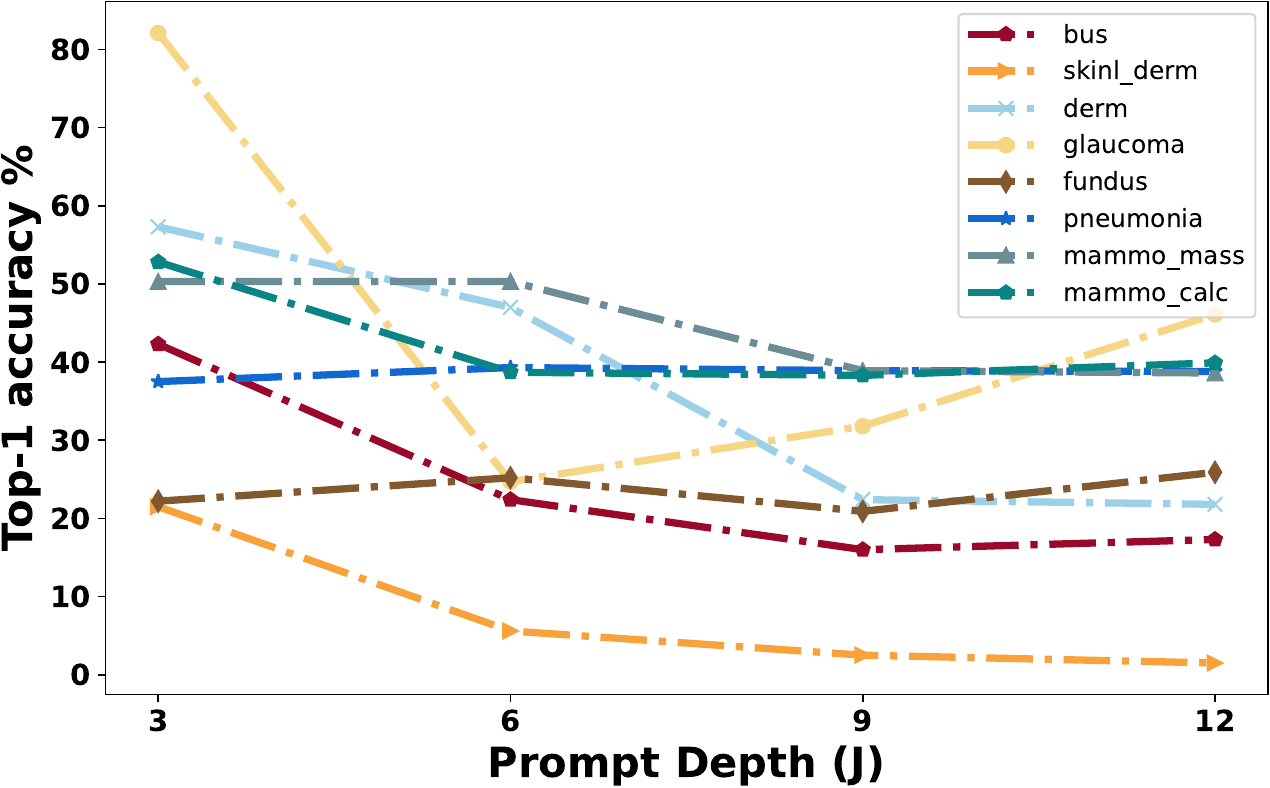}
        \label{fig:prompt_depth}
    \end{minipage}\hfill
    \vspace{-5mm}
    % \captionsetup{font=scriptsize}
    \caption {\small \textbf{(left)} Influence of the number of training shots (1, 2, 4, 8, 16), showing that additional shots improve performance for some datasets while others remain stable. \textbf{(right)} Impact of varying prompt depth (J) on Top-1 accuracy across datasets, highlighting performance degradation with increasing depth.}
    \label{fig:ablation_2}
    \vspace{-5mm}
\end{figure*}

\noindent\textbf{e. Effect of loss components:}
We evaluate the impact of different loss components on the mammo-mass and glaucoma datasets. Using only cross-entropy loss ($L_{ce}$) achieves a baseline accuracy of 50.5\% for mammo-mass and 82.1\% for glaucoma, while adding KL divergence regularization ($L_{kl}$) improves performance to 50.8\% and 82.7\%, respectively. Furthermore, to assess cross-dataset generalization, we train on one dataset and test on related datasets. For mammo-mass, transferring to mammo-calc maintains stable performance (53.4\%) across all configurations. However, on bus, performance declines when adding contrastive loss, indicating that contrastive learning may not always enhance generalization.

% Our cross-dataset and domain generalization experiments confirm that our decoupled multi-modal framework significantly bolsters out-of-distribution (OOD) robustness in medical imaging. By separating clinically relevant, invariant features from spurious correlations in both vision and text modalities, the method outperforms standard single-modality or holistic multi-modal baselines. Notably, training on one chest X-ray dataset and testing on another highlights its ability to retain diagnostic signals despite changes in patient populations or imaging protocols—vital for real-world healthcare settings. Moreover, domain generalization across MedIMeta subsets shows marked resilience to unseen data distributions, underscoring the value of controlled feature interactions in vision and language. The synergy between prompt-based adaptation and decoupling also accelerates few-shot learning, which is especially beneficial for medical domains where annotated data are often scarce.

\vspace{-5mm}
\section{Conclusion}
\vspace{-5mm}
We propose a structured decoupling framework for medical vision-language models that improves out-of-distribution generalization by isolating invariant clinical features from task-agnostic correlations in both image and text. Using LoRA-based fine-tuning and learnable prompts, DRiFt enhances cross-modal alignment with minimal overhead. By disentangling features at both visual and textual levels, the model focuses on clinically relevant cues while suppressing noise, outperforming existing multi-modal prompt-tuning approaches. This work underscores the importance of targeted loss design and structured feature separation for robust generalization in medical AI.

\vspace{-2mm}

\begin{credits}
\subsubsection{\discintname}
The authors have no competing interests to declare that are
relevant to the content of this article.
\end{credits}
\bibliographystyle{unsrt}
\bibliography{ref}

\begin{thebibliography}{10}

\bibitem{sintchenko2003clinical}
Vitali Sintchenko and Enrico~W Coiera.
\newblock Which clinical decisions benefit from automation? a task complexity approach.
\newblock {\em International journal of medical informatics}, 70(2-3):309--316, 2003.

\bibitem{rahman2024mrishift}
Umaima Rahman, Guangyi Chen, and Kun Zhang.
\newblock Mrishift: Disentangled representation learning for 3d mri lesion segmentation under distributional shifts.
\newblock In {\em 2024 12th European Workshop on Visual Information Processing (EUVIP)}, pages 1--6. IEEE, 2024.

\bibitem{white2024heterogeneity}
Samuel~J White, Qi~Sheng Phua, Lucy Lu, Kaspar~L Yaxley, Matthew~DF Mcinnes, and Minh-Son To.
\newblock Heterogeneity in systematic reviews of medical imaging diagnostic test accuracy studies: a systematic review.
\newblock {\em JAMA Network Open}, 7(2):e240649--e240649, 2024.

\bibitem{saab2022reducing}
Khaled Saab, Sarah Hooper, Mayee Chen, Michael Zhang, Daniel Rubin, and Christopher R{\'e}.
\newblock Reducing reliance on spurious features in medical image classification with spatial specificity.
\newblock In {\em Machine Learning for Healthcare Conference}, pages 760--784. PMLR, 2022.

\bibitem{compton2023more}
Rhys Compton, Lily Zhang, Aahlad Puli, and Rajesh Ranganath.
\newblock When more is less: Incorporating additional datasets can hurt performance by introducing spurious correlations.
\newblock In {\em Machine Learning for Healthcare Conference}, pages 110--127. PMLR, 2023.

\bibitem{khattak2023maple}
Muhammad~Uzair Khattak, Hanoona Rasheed, Muhammad Maaz, Salman Khan, and Fahad~Shahbaz Khan.
\newblock Maple: Multi-modal prompt learning.
\newblock In {\em Proceedings of the IEEE/CVF conference on computer vision and pattern recognition}, pages 19113--19122, 2023.

\bibitem{zhou2022conditional}
Kaiyang Zhou, Jingkang Yang, Chen~Change Loy, and Ziwei Liu.
\newblock Conditional prompt learning for vision-language models.
\newblock In {\em Proceedings of the IEEE/CVF conference on computer vision and pattern recognition}, pages 16816--16825, 2022.

\bibitem{rahman2025can}
Umaima Rahman, Raza Imam, Mohammad Yaqub, Boulbaba~Ben Amor, and Dwarikanath Mahapatra.
\newblock Can language-guided unsupervised adaptation improve medical image classification using unpaired images and texts?
\newblock In {\em 2025 IEEE 22nd International Symposium on Biomedical Imaging (ISBI)}, pages 1--5. IEEE, 2025.

\bibitem{rahman2025dimple}
Umaima Rahman, Mohammad Yaqub, and Dwarikanath Mahapatra.
\newblock Dimple--disentangled multi-modal prompt learning: Enhancing out-of-distribution alignment with invariant and spurious feature separation.
\newblock {\em arXiv preprint arXiv:2506.21237}, 2025.

\bibitem{hu2022lora}
Edward~J Hu, Yelong Shen, Phillip Wallis, Zeyuan Allen-Zhu, Yuanzhi Li, Shean Wang, Lu~Wang, Weizhu Chen, et~al.
\newblock Lora: Low-rank adaptation of large language models.
\newblock {\em ICLR}, 1(2):3, 2022.

\bibitem{imam2024test}
Raza Imam, Hanan Gani, Muhammad Huzaifa, and Karthik Nandakumar.
\newblock Test-time low rank adaptation via confidence maximization for zero-shot generalization of vision-language models.
\newblock {\em arXiv preprint arXiv:2407.15913}, 2024.

\bibitem{woerner2024comprehensive}
Stefano Woerner, Arthur Jaques, and Christian~F Baumgartner.
\newblock A comprehensive and easy-to-use multi-domain multi-task medical imaging meta-dataset (medimeta).
\newblock {\em arXiv preprint arXiv:2404.16000}, 2024.

\bibitem{dai2023instructblipgeneralpurposevisionlanguagemodels}
Wenliang Dai, Junnan Li, Dongxu Li, Anthony Meng~Huat Tiong, Junqi Zhao, Weisheng Wang, Boyang Li, Pascale Fung, and Steven Hoi.
\newblock Instructblip: Towards general-purpose vision-language models with instruction tuning, 2023.

\bibitem{li2022domain}
Chenxin Li, Xin Lin, Yijin Mao, Wei Lin, Qi~Qi, Xinghao Ding, Yue Huang, Dong Liang, and Yizhou Yu.
\newblock Domain generalization on medical imaging classification using episodic training with task augmentation.
\newblock {\em Computers in biology and medicine}, 141:105144, 2022.

\bibitem{yoon2024domain}
Jee~Seok Yoon, Kwanseok Oh, Yooseung Shin, Maciej~A Mazurowski, and Heung-Il Suk.
\newblock Domain generalization for medical image analysis: A review.
\newblock {\em Proceedings of the IEEE}, 2024.

\bibitem{guo2022evaluation}
Lin~Lawrence Guo, Stephen~R Pfohl, Jason Fries, Alistair~EW Johnson, Jose Posada, Catherine Aftandilian, Nigam Shah, and Lillian Sung.
\newblock Evaluation of domain generalization and adaptation on improving model robustness to temporal dataset shift in clinical medicine.
\newblock {\em Scientific reports}, 12(1):2726, 2022.

\bibitem{shakeri2024few}
Fereshteh Shakeri, Yunshi Huang, Julio Silva-Rodr{\'\i}guez, Houda Bahig, An~Tang, Jose Dolz, and Ismail Ben~Ayed.
\newblock Few-shot adaptation of medical vision-language models.
\newblock In {\em International Conference on Medical Image Computing and Computer-Assisted Intervention}, pages 553--563. Springer, 2024.

\bibitem{lai2023clipath}
Zhengfeng Lai, Zhuoheng Li, Luca~Cerny Oliveira, Joohi Chauhan, Brittany~N Dugger, and Chen-Nee Chuah.
\newblock Clipath: Fine-tune clip with visual feature fusion for pathology image analysis towards minimizing data collection efforts.
\newblock In {\em Proceedings of the IEEE/CVF International Conference on Computer Vision}, pages 2374--2380, 2023.

\bibitem{imam2025noise}
Raza Imam, Asif Hanif, Jian Zhang, Khaled~Waleed Dawoud, Yova Kementchedjhieva, and Mohammad Yaqub.
\newblock Noise is an efficient learner for zero-shot vision-language models.
\newblock {\em arXiv preprint arXiv:2502.06019}, 2025.

\bibitem{rahman2024meduna}
Umaima Rahman, Raza Imam, Dwarikanath Mahapatra, and Boulbaba~Ben Amor.
\newblock Meduna: Language guided unsupervised adaptation of vision-language models for medical image classification.
\newblock {\em arXiv preprint arXiv:2409.02729}, 2024.

\bibitem{liu2022learning}
Xiao Liu, Pedro Sanchez, Spyridon Thermos, Alison~Q O’Neil, and Sotirios~A Tsaftaris.
\newblock Learning disentangled representations in the imaging domain.
\newblock {\em Medical Image Analysis}, 80:102516, 2022.

\bibitem{sarafraz2024domain}
Gita Sarafraz, Armin Behnamnia, Mehran Hosseinzadeh, Ali Balapour, Amin Meghrazi, and Hamid~R Rabiee.
\newblock Domain adaptation and generalization of functional medical data: A systematic survey of brain data.
\newblock {\em ACM Computing Surveys}, 56(10):1--39, 2024.

\bibitem{shenzhenTB}
{Shenzhen Hospital}.
\newblock Shenzhen hospital chest x-ray (cxr) set.
\newblock \href{}{https://data.lhncbc.nlm.nih.gov/public/Tuberculosis-Chest-X-ray-Datasets/Shenzhen-Hospital-CXR-Set/index.html}.

\bibitem{montgomeryTB}
{Montgomery County}.
\newblock Montgomery county chest x-ray (cxr) set.
\newblock \href{}{https://data.lhncbc.nlm.nih.gov/public/Tuberculosis-Chest-X-ray-Datasets/Montgomery-County-CXR-Set/MontgomerySet/index.html}.

\bibitem{porwal2018indian}
Prasanna Porwal, Samiksha Pachade, Ravi Kamble, Manesh Kokare, Girish Deshmukh, Vivek Sahasrabuddhe, and Fabrice Meriaudeau.
\newblock Indian diabetic retinopathy image dataset (idrid): a database for diabetic retinopathy screening research.
\newblock {\em Data}, 3(3):25, 2018.

\bibitem{isicArchive}
{International Skin Imaging Collaboration}.
\newblock Isic archive: International skin imaging collaboration dataset.
\newblock \href{}{https://www.isic-archive.com/}.

\bibitem{imam2025robustness}
Raza Imam, Rufael Marew, and Mohammad Yaqub.
\newblock On the robustness of medical vision-language models: Are they truly generalizable?
\newblock {\em arXiv preprint arXiv:2505.15425}, 2025.

\bibitem{zhangamend}
Jie Zhang, Xiaosong Ma, Song Guo, Peng Li, Wenchao Xu, Xueyang Tang, and Zicong Hong.
\newblock Amend to alignment: Decoupled prompt tuning for mitigating spurious correlation in vision-language models.
\newblock In {\em Forty-first International Conference on Machine Learning}.

\end{thebibliography}
\end{document}